\documentclass{article}
\usepackage{spconf,amsmath,graphicx}
\usepackage{multirow}

\title{An efficient deep learning hashing neural network for mobile visual search}
\name{Heng Qi, Wu Liu, Liang Liu \thanks{This work was partially supported by the CCF-Tencent Open Research Fund (No. AGR20160113), the National Natural Science Foundation of China (No. 61632008), and the Fundamental Research Funds for the Central Universities (No. 2016RCGD32).}}
\address{
Beijing Key Laboratory of Intelligent Telecommunication Software and Multimedia, \\
Beijing University of Posts and Telecommunications, Beijing 100876, China\\
\{qiheng, liuwu, liangliu\}@bupt.edu.cn}
\begin{document}
%
\maketitle
\begin{abstract}
Mobile visual search applications are emerging that enable users to sense their surroundings with smart phones. However, because of the particular challenges of mobile visual search, achieving a high recognition bitrate has becomes a consistent target of previous related works. In this paper, we propose a few-parameter, low-latency, and high-accuracy deep hashing approach for constructing binary hash codes for mobile visual search. First, we exploit the architecture of the MobileNet model, which significantly decreases the latency of deep feature extraction by reducing the number of model parameters while maintaining accuracy. Second, we add a hash-like layer into MobileNet to train the model on labeled mobile visual data. Evaluations show that the proposed system can exceed state-of-the-art accuracy performance in terms of the MAP. More importantly, the memory consumption is much less than that of other deep learning models. The proposed method requires only $13$ MB of memory for the neural network and achieves a MAP of $97.80\%$ on the mobile location recognition dataset used for testing.
\end{abstract}
\begin{figure}[htb]

  \centering
  \centerline{\includegraphics[width=8.5cm]{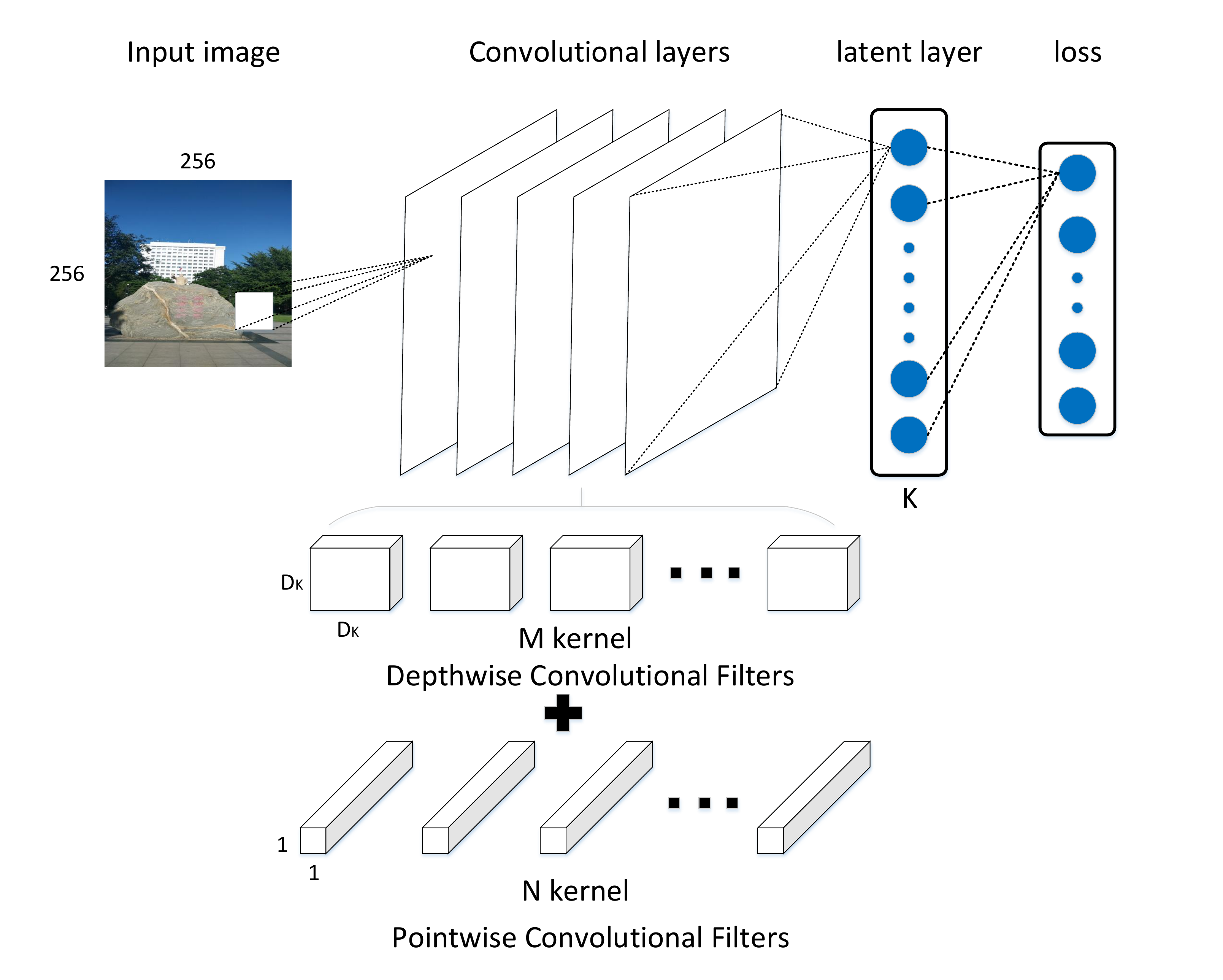}}
%
\caption{The architecture of the proposed deep learning hashing neural network for mobile visual search}
\label{fig:res}
\end{figure}
\section{Introduction}
\label{sec:intro}

With the proliferation of mobile devices, it is becoming possible to use mobile perception functionalities (e.g., cameras, GPS, and Wi-Fi) to perceive the surrounding environment \cite{DBLP:journals/ejivp/LiuMQZC17}. Among such techniques, mobile visual search plays a key role in mobile localization, mobile media search, and mobile social networking. However, rather than simply porting traditional visual search methods to mobile platforms, for mobile visual search, one must face the challenges of a large aural-visual variance of queries, stringent memory and computation constraints, network bandwidth limitations, and the desire for an instantaneous search experience.

Most research on mobile visual search has predominantly focused on achieving high recognition bitrates \cite{DBLP:journals/tmm/LiuMZ14, DBLP:journals/ijcv/ChandrasekharTCTRGG12, DBLP:conf/cvpr/HeFLCLCC12}. Recently, an increasing number of researchers are attempting to exploit feature signatures produced through hashing in mobile visual search because of the good balance that can be achieved among computation and memory requirements, training efficiency, quantization complexity, and search performance. However, most of the existing hashing-based mobile visual search methods attempt to compress existing classical handcrafted features into binary code. These methods all focus on how to decrease the loss suffered during compression. Only a few of them attempt to automatically learn effective binary code features from a large-scale image dataset using a deep neural network. There are two main reasons for this: 1) the lack of effective deep learning hashing methods for mobile visual search and 2) the high computational complexity of existing deep neural networks.

For traditional visual search, convolutional neural networks have become ubiquitous \cite{DBLP:journals/ijcv/RussakovskyDSKS15}\cite{ DBLP:conf/cvpr/LiuMZCL15}. Studies have shown that the deep features learned using such networks capture rich image representations and enable better performance than handcrafted features in visual classification \cite{DBLP:conf/aaai/XiaPLLY14}\cite{DBLP:conf/cvpr/GanYYYM16}\cite{DBLP:journals/ijcv/GanYZZZ16} \cite{gan2016concepts}, object detection \cite{DBLP:conf/cvpr/CiresanMS12}\cite{DBLP:conf/cvpr/GirshickDDM14}, semantic segmentation \cite{DBLP:journals/corr/SermanetEZMFL13}, and image retrieval \cite{DBLP:conf/nips/KrizhevskySH12}\cite{chu_tmm2013}. The general trend in research on deep learning methods is to construct deeper and more complicated networks to achieve higher accuracy \cite{DBLP:journals/corr/SimonyanZ14a}\cite{DBLP:journals/corr/SzegedyVISW15}\cite{DBLP:conf/cvpr/GanWYYH15}. In the field of mobile visual search, however, a deeper neural network model consumes more memory and time, which cannot be easily supplied by mobile devices.
To adapt deeper neural networks for mobile devices, a new network architecture called MobileNet has recently been proposed by Google \cite{DBLP:journals/corr/HowardZCKWWAA17}. In the MobileNet model, a standard convolutional layer is decomposed into a depthwise convolutional layer and a pointwise convolutional layer, thereby greatly reducing the amount of calculation necessary and the model size. This model works very well on image classification problems.

Inspired by the works presented in \cite{DBLP:journals/ejivp/LiuMQZC17} and \cite{DBLP:journals/corr/HowardZCKWWAA17}, to develop a more effective and efficient mobile visual search system, this paper proposes to combine a few-parameter, low-latency, and high-accuracy network architecture with a hash function. We incorporate the hash function as a latent layer between the image representations and the classification outputs in MobileNet,
which allows us not only to maintain the accuracy of the visual search but also to adapt the model to the mobile environment. An overview of the system is illustrated in Figure 1. The final binary hash codes can be learned by minimizing an objective function defined over the classification error.
Experimental results on a mobile location recognition dataset show that our method achieves superior performance compared with other hashing approaches.

\section{Architecture}
\label{sec:arch}

In this section, we first describe the MobileNet network structure and the convolution layer structure, which is based on depthwise separable filters and pointwise separable filters. Then, we describe the hash layer in our convolutional neural network and train a convolutional neural network model that exploits semantic labels to automatically create binary codes.

\begin{table}
\caption{The MobileNet-based hashing network for mobile visual search}
\footnotesize
\begin{center}
\begin{tabular}{p{2cm}|p{3cm}|p{2.5cm}}
\hline
Type / Stride & Filter Shape  & Input Size\\
\hline
Conv / s2  & 3 $\times$ 3 $\times$ 3 $\times$ 32 & 224 $\times$ 224 $\times$ 3\\
\hline
Conv dw / s1  & 3 $\times$ 3 $\times$ 32 dw & 112 $\times$ 112 $\times$ 32\\
\hline
Conv / s1  & 1 $\times$ 1 $\times$ 32 $\times$ 64 & 112 $\times$ 112 $\times$ 32\\
\hline
Conv dw / s2  & 3 $\times$ 3 $\times$ 64 dw & 112 $\times$ 112 $\times$ 64\\
\hline
Conv / s1  & 1 $\times$ 1 $\times$ 64 $\times$ 128 & 56 $\times$ 56 $\times$ 64\\
\hline
Conv dw / s1  & 3 $\times$ 3 $\times$ 128 dw & 56 $\times$ 56 $\times$ 128\\ \hline
Conv / s1  & 1 $\times$ 1 $\times$ 128 $\times$ 128 & 56 $\times$ 56 $\times$ 128\\ \hline
Conv dw / s2  & 3 $\times$ 3 $\times$ 128 dw & 56 $\times$ 56 $\times$ 128\\ \hline
Conv / s1  & 1 $\times$ 1 $\times$ 128 $\times$ 256 & 28 $\times$ 28 $\times$ 128\\ \hline
Conv dw / s1  & 3 $\times$ 3 $\times$ 256 dw & 28 $\times$ 28 $\times$ 256\\ \hline
Conv / s1  & 1 $\times$ 1 $\times$ 256 $\times$ 256 & 28 $\times$ 28 $\times$ 256\\ \hline
Conv dw/ s2  & 3 $\times$ 3 $\times$ 256 dw & 28 $\times$ 28 $\times$ 256\\ \hline
Conv / s1  & 1 $\times$ 1 $\times$ 256 $\times$ 512 & 14 $\times$ 14 $\times$ 256\\ \hline

Conv dw / s1  & 3 $\times$ 3 $\times$ 512 dw & 14 $\times$ 14 $\times$ 512\\
Conv / s1  & 1 $\times$ 1 $\times$ 512 $\times$ 512 & 14 $\times$ 14 $\times$ 512\\ \hline

Conv dw / s2  & 3 $\times$ 3 $\times$ 512 dw & 14 $\times$ 14 $\times$ 512\\ \hline
Conv / s1  & 1 $\times$ 1 $\times$ 512 $\times$ 1024 & 7 $\times$ 7 $\times$ 512\\ \hline
Conv dw/ s2  & 3 $\times$ 3 $\times$ 1024 dw & 7 $\times$ 7 $\times$ 1024\\ \hline
Conv / s1  & 1 $\times$ 1 $\times$ 1024 $\times$ 1024 & 7 $\times$ 7 $\times$ 1024\\ \hline
Avg Pool / s1  & Pool 7 $\times$ 7 & 7 $\times$ 7 $\times$ 1024\\ \hline
FC / s1  & 1024 $\times$ 64 & 1 $\times$ 1 $\times$ 64\\ \hline
Sigmoid  & In place & 1 $\times$ 1 $\times$ 64\\ \hline
FC / s1  & 1024 $\times$ 162 & 1 $\times$ 1 $\times$ 162\\ \hline
Softmax  & Classifier & 1 $\times$ 1 $\times$ 162\\ \hline
\end{tabular}
\end{center}
\vspace{-5mm}
\end{table}
\subsection{MobileNet}
\label{ssec:mobilenet}

The MobileNet model is built on depthwise separable convolutions. These convolutions factorize a standard convolution into a depthwise convolution and a $1 \times 1$ convolution called a pointwise convolution. The MobileNet architecture is defined in Table 1. In MobileNet, the depthwise convolution involves the application of a single filter to each input channel. The pointwise convolution then involves the application of a $1 \times 1$ convolution to combine the outputs of the depthwise convolution. A standard convolutional layer takes as input a $D_{F} \times D_{F} \times M$ feature map F and produces a $D_{F} \times D_{F} \times N$ feature map G, where $D_{F}$ is the spatial width and height of the square input feature map, $M$ is the number of input channels (input depth), $D_G$ is the spatial width and height of the square output feature map, and $N$ is the number of output channel.

A standard convolution has the following computational cost:
\begin{equation}D_K \cdot D_K \cdot M \cdot N \cdot D_F \cdot D_F\end{equation}
The corresponding cost of depthwise separable convolutions is as follows:
\begin{equation}D_K \cdot D_K \cdot M \cdot D_F \cdot D_F + M \cdot N \cdot D_F \cdot D_F\end{equation}
which is the sum of the costs of the depthwise and $1 \times 1$ pointwise convolutions. By expressing a standard convolution as a two-step process of filtering and combining, we achieve the following reduction in computational cost:
\begin{equation}\frac{D_K \cdot D_K \cdot M \cdot D_F \cdot D_F + M \cdot N \cdot D_F \cdot D_F}{D_K \cdot D_K \cdot M \cdot N \cdot D_F \cdot D_F}\end{equation}

MobileNet uses $3 \times 3$ depthwise separable convolutions, which require 8$\sim$9 times less computation than standard convolutions with only a small reduction in accuracy.

Table 2 compares MobileNet with other popular models in terms of accuracy as assessed on the ImageNet database, number of multi-adds (million), and number of parameters (million). MobileNet is nearly as accurate as VGG-16 \cite{DBLP:journals/corr/SimonyanZ14a} while being 32 times smaller and 27 times less compute intensive. It is more accurate than GoogleNet \cite{DBLP:conf/cvpr/SzegedyLJSRAEVR15} while being smaller and requiring more than 2.5 times less computation. MobileNet \cite{DBLP:journals/corr/HowardZCKWWAA17} is also more accurate than Squeezenet \cite{DBLP:journals/corr/IandolaMAHDK16} and AlexNet \cite{DBLP:conf/nips/KrizhevskySH12} with a nearly identical model size and less computation.

\begin{table}
\caption{Comparisons between MobileNet and other popular deep learning models \cite{DBLP:journals/corr/HowardZCKWWAA17}.}
\begin{center}
\begin{tabular}{l|l|l|l}
\hline
\footnotesize{Model} & \footnotesize{Accuracy}  & \footnotesize{Multi-Adds ($10^6$)} & \footnotesize{Parameters ($10^6$)}\\
\hline
AlexNet  & 57.2\% & 720 & 60\\
Squeezenet  & 57.5\% & 1700 & 1.25\\
GoogleNet  & 69.8\% & 1550 & 6.8\\
VGG-16  & 71.5\% & 15300 & 138\\
MobileNet  & 70.6\% & 569 & 4.2\\
\hline
\end{tabular}
\end{center}
\vspace{-8mm}
\end{table}

\vspace{-4mm}

\subsection{Hash Function}
\label{ssec:hashfunc}

Hash mapping, which is required for learning from images, is based on the following principles. The hash codes should respect the semantic similarity between image labels. Images that share the same class labels should be mapped to similar binary codes. Let $I = \{I_n\}^{N}_{n=1}$ denote $N$ images, and let $Y = {y_n}^{M \times N}$ be their associated label vectors, where $M$ is the total number of class labels. An entry in $y_n$ has a value of 1 if the image $I_n$ belongs to the corresponding class and is 0 otherwise. Our goal is to learn a mapping $F : I \to \{0, 1\}^{K \times N}$ that maps the images to their $k$-bit binary codes $B = {b_n} \in \{0, 1\}^{K \times N}$ while preserving the semantic similarity relationships among the image data \cite{DBLP:journals/corr/YangLC15a}.

Our network is built on MobileNet. Each layer is followed by a batchnorm layer and a nonlinear ReLU layer, with the exception of the fully connected layer for classification. A final average pooling layer reduces the spatial resolution to 1 before the fully connected layer. To incorporate the learned features into the binary codes, we add a latent layer $H$ with $K$ units to the top of layer $pool6$, as illustrated in Figure 1. This latent layer is fully connected to $pool6$ and uses sigmoid units so that the activations are bounded between 0 and 1 \cite{DBLP:journals/jmlr/GlorotBB11}. Let $W_H \in R^{d \times K}$ denote the weights (i.e., the projection matrix) in the latent layer. For a given image $I_n$ with the feature vector $a_6 \in R^d$ in layer $pool6$, the activations of the units in $H$ can be computed as $a^H_n = \sigma(a^6_n \cdot W^H + e^H)$. Here, $e^H$ is the bias term, and $\sigma( \cdot )$ is the logistic sigmoid function, which is defined as $\sigma(z) = 1/(1 + exp(-z))$, where $z$ is a real value. The binary encoding function is given by
\begin{equation}
b_n = sign(\sigma(a^6_n \cdot W^H + e^H) - 0.5) = sign(a^H_n - 0.5)
\end{equation}
where $sign(v) = 1$ if $v > 0$ and $sign(v) = 0$ otherwise. $sign(\cdot)$ performs element-wise operations on a matrix or a vector.

\section{Experiments}

\begin{table}
\caption{The MAPs of different hashing methods for mobile location recognition.}
\begin{center}
\begin{tabular}{p{2cm}|p{1.5cm}|p{1.5cm}|p{1.5cm}}
\hline
\multirow{2}* {Method} & \multicolumn{3}{c}{MAP}\\
\cline{2-4} & 16-bit & 32-bit & 64-bit\\
\hline
VHB \cite{DBLP:conf/cvpr/HeFLCLCC12} & - & - & 19.36\\
\hline
SSFS \cite{DBLP:conf/pcm/WangZMX16} & - & - & 20.22\\
\hline
DLBH \cite{DBLP:conf/cvpr/LinYHC15}  & 59.80 & 78.68 & 87.15\\
\hline
SSDH \cite{DBLP:journals/corr/YangLC15a}  & 78.26 & 91.82 & 92.43\\
\hline
\textbf{Our method}  & \textbf{96.66} & \textbf{97.61} & \textbf{97.80}\\
\hline
\end{tabular}
\end{center}
\vspace{-5mm}
\end{table}

To evaluate the proposed method, we conducted experiments on the mobile location recognition dataset presented in \cite{DBLP:conf/pcm/WangZMX16}. The dataset contains 8,062 images, which were captured from 162 locations. We used the same experimental parameters as in \cite{DBLP:conf/pcm/WangZMX16} and \cite{DBLP:journals/ejivp/LiuMQZC17}.
We implemented our approach using the open-source CAFFE \cite{DBLP:conf/mm/JiaSDKLGGD14} package. We initialized the network parameters by adopting the parameters of a MobileNet trained on the 14 million images of the ImageNet \cite{DBLP:conf/nips/KrizhevskySH12} dataset. The parameters in the hash layer were randomly initialized.
The learning rate was initialized as 0.01 and was decreased to 1/10 of its previous value after every 10,000 iterations. The entire training procedure terminated after 30,000 iterations. For the learning of the network parameters, in conjunction with backpropagation, we exploited the mini-batch stochastic gradient descent algorithm with a mini-batch size of 32 images to minimize the classification error. Our model is a lightweight modification of MobileNet and thus is easy to implement.

In the evaluation, we used the Mean Average Precision (MAP) as the evaluation criterion. We ranked all of the images according to their Hamming distances from the query image, selected the top k images from the ranked list as the retrieval results, and computed the MAP on these retrieved images. We set k to 100 in these experiments.
We used the class labels as the ground truth and adopted the common settings for computing the MAP by examining whether the retrieved images and the query shared common class labels.
We compared our method with Visual Hash Bits (VHB) \cite{DBLP:conf/cvpr/HeFLCLCC12} and Space-Saliency Fingerprint Selection-based hash codes (SSFS) \cite{DBLP:conf/pcm/WangZMX16}, which are both traditional hashing methods for mobile location recognition. The other two methods considered for comparison, Deep Learning of Binary Hash Codes (DLBH) \cite{DBLP:conf/cvpr/LinYHC15} and Supervised Semantics-Preserving Deep Hashing (SSDH) \cite{DBLP:journals/corr/YangLC15a}, are both deep-learning-based hashing methods.

Table 3 compares the performances on the dataset of the different hashing methods for different hash code lengths. From the results, we can see that the accuracies of the deep learning hashing methods greatly exceed those of the traditional hashing methods, which demonstrates the power of deep learning technology for binary code learning. Furthermore, because SSDH learns the feature representations and binary codes simultaneously and imposes more constraints for binary code learning, it achieves higher performance than DLBH. Our method achieves the best results on this dataset for all of the different hash code lengths. Especially when the hash code length is relatively short, our method can achieve extremely high accuracy. This can be attributed to the fact that MobileNet networks enable the joint learning of representations and hash functions from images. Moreover, the learned representations are more effective and stable than those of AlexNet-based models such as DLBH and SSDH. Finally, the sizes of the DLBH and SSDH models are greater than 230 MB, whereas for MobileNet, the model size is only 13 MB. Thus, the memory demand for a mobile device is reduced by a factor of approximately 18.

\section{CONCLUSION}
\label{sec:conclusion}

We present an effective deep learning framework based on MobileNet for creating hash-like binary codes for mobile visual search.
In this framework, we incorporate the hash function as a latent layer between the feature layer and the output layer in the MobileNet network. By optimizing an objective function defined over the classification error, our method jointly learns the binary codes, features, and classification results. To evaluate the performance of the proposed network,
we applied it to a mobile location recognition dataset. The experimental results demonstrate that it can achieve a MAP improvement of $5.5\%$ compared with state-of-the-art methods. The model requires only 13 MB of memory.


\end{document}